%% file: main.tex
\input{_constants}

\cameraready

\pdfoutput=1
\documentclass[10pt,twocolumn,letterpaper]{article}
\usepackage{authblk}
 
\input{cvpr_header}

\begin{document}

\title{\emph{ProMotion}: Prototypes As Motion Learners}
\authorBlock

\twocolumn[{%
\renewcommand\twocolumn[1][]{#1}%
\maketitle
\begin{center}
    \centering
    \captionsetup{type=figure}
    \includegraphics[width=.88\textwidth,height=6.5cm]{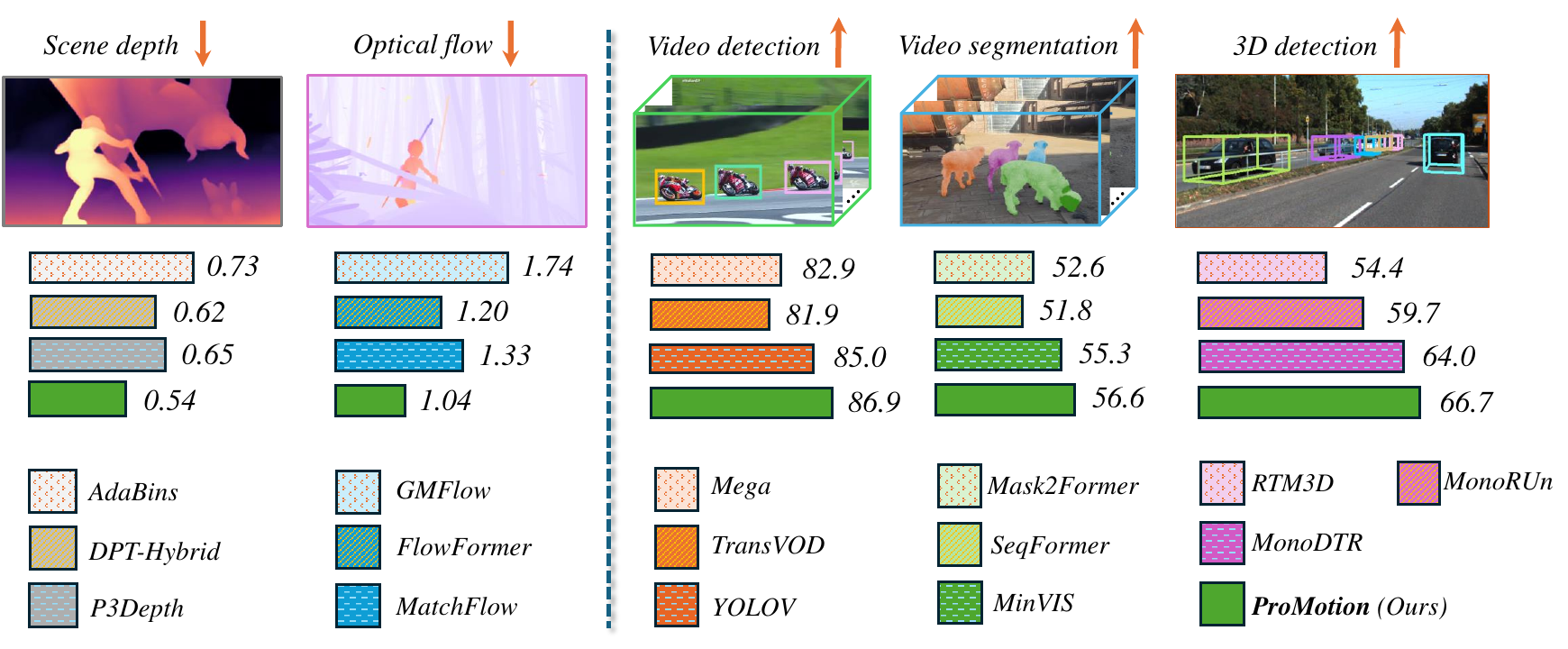}
    \put(-385,42){\small{~\cite{bhat2021adabins}}}  
    \put(-372,25){\small{~\cite{ranftl2021vision}}}  
    \put(-383,9){\small{~\cite{P3Depth}}}  
        \put(-299,43){\small{~\cite{xu2022gmflow}}}  
        \put(-288,26){\small{~\cite{huang2022flowformer}}}  
        \put(-292,10){\small{~\cite{dong2023rethinking}}}  
            \put(-220,44){\small{~\cite{chen20mega}}} 
            \put(-206,28){\small{~\cite{zhou2022transvod}}}  
            \put(-213,10){\small{~\cite{shi2023yolov}}}  
                \put(-120,44){\small{~\cite{cheng2021mask2former}}} 
                \put(-128,28){\small{~\cite{wu2022seqformer}}} 
                \put(-137,11){\small{~\cite{huang2022minvis}}}  
                    \put(-66.8,44){\small{~\cite{li2020rtm3d}}} 
                    \put(-57,28){\small{~\cite{huang2022monodtr}}}  
                    \put(-4,46){\small{~\cite{monorun2021}}}  
    \vspace{-3.5mm}
    \captionof{figure}{\textbf{\textsc{ProMotion} harmoniously handle core motion tasks (\ie, optical flow and scene depth estimation), within an elegant prototype-based framework.} This adaptability further enhances its utility in enabling seamless knowledge transfer to various downstream applications. For depth~\cite{bhat2021adabins, ranftl2021vision, P3Depth}, we averagely reduce 0.13 $Abs\ Rel$; For flow~\cite{xu2022gmflow, huang2022flowformer, dong2023rethinking}, we averagely reduce 0.38 \textit{AEPE}; For downstream tasks, we averagely achieve 3.6\% $mAP$ boost in video object detection~\cite{chen20mega, zhou2022transvod, shi2023yolov}, 3.4\% $AP$ boost in video object segmentation~\cite{cheng2021mask2former, wu2022seqformer, huang2022minvis}, and 7.3\% $AP_{3D}$ boost in 3D object detection~\cite{li2020rtm3d, huang2022monodtr, monorun2021}.} 
    \label{fig:intro_fig}
\end{center}%
}]

\maketitle
\input{00_abstract}

\input{01_intro}

\input{02_related}

\input{03_method}

\input{10_conclusion}

\vspace{-3mm}
\section*{Acknowledgement}
\vspace{-2mm}
This work is supported by the National Science Foundation under Award No. 2242243.

{\small
\bibliographystyle{ieeenat_fullname}
\bibliography{11_references}
}

\end{document}

%% file: _constants.tex
\def\paperID{4220}
\def\confName{CVPR}
\def\confYear{2024}

\def\authorBlock{
  \author[1]{Yawen Lu}
  \author[2*]{Dongfang Liu}
  \author[3]{Qifan Wang}
  \author[2]{Cheng Han}
  \author[4]{Yiming Cui}
  \author[1]{Zhiwen Cao}
  \author[2]{Xueling Zhang}
  \author[1]{Yingjie Victor Chen} 
  \author[6*]{Heng Fan}
  \affil[1]{Purdue University}
  \affil[2]{Rochester Institute of Technology}
  \affil[3]{Meta AI}
  \affil[4]{University of Florida}
  \affil[6]{University of North Texas}  

}

\newif\ifreview 
\newif\ifarxiv 
\newif\ifcamera \newcommand{\cameraready}{\cameratrue}
\newif\ifrebuttal

%% file: cvpr_header.tex
\ifreview \usepackage[review]{cvpr} \fi
\ifarxiv \usepackage[pagenumbers]{cvpr} \fi
\ifrebuttal \usepackage[rebuttal]{cvpr} \fi
\ifcamera \usepackage{cvpr} \fi
\usepackage{cvpr}

\input{_macros}  

\usepackage{xr-hyper}

\makeatletter
\newcommand*{\addFileDependency}[1]{
  \typeout{(#1)}
  \@addtofilelist{#1}
  \IfFileExists{#1}{}{\typeout{No file #1.}}
}

\makeatother

\definecolor{cvprblue}{rgb}{0.21,0.49,0.74}
\usepackage[pagebackref,breaklinks,colorlinks,citecolor=cvprblue]{hyperref}
\usepackage[capitalize]{cleveref}
\crefname{section}{Sec.}{Secs.}
\crefname{table}{Table}{Tables}
\crefname{figure}{Fig.}{Figs.}

%% file: _macros.tex
\usepackage{graphicx}	
\usepackage{amsmath}	
\usepackage{amssymb}	
\usepackage{booktabs}
\usepackage{times}
\usepackage{microtype}
\usepackage{epsfig}
\usepackage[table,xcdraw,dvipsnames]{xcolor}
\usepackage{arydshln}
\usepackage{placeins}
\usepackage{color, colortbl}
\usepackage{stfloats}
\usepackage{enumitem}
\usepackage{tabularx}
\usepackage{xstring}
\usepackage{multirow}
\usepackage{xspace}
\usepackage{url}
\usepackage{xcolor}
\usepackage[hang,flushmargin]{footmisc}

\usepackage{graphicx}
\usepackage{amsmath}
\usepackage{amssymb}
\usepackage{booktabs}
\usepackage{array}
\usepackage{multirow}
\usepackage{makecell}
\usepackage{bbding}
\PassOptionsToPackage{normalem}{ulem}
\usepackage{ulem}
\usepackage{arydshln}
\usepackage{xcolor}         
\usepackage[accsupp]{axessibility}
\usepackage{pifont}
\usepackage{combelow} 

\usepackage{graphicx}
\usepackage{amssymb}
\usepackage{xcolor}
\usepackage{pifont}
\usepackage{multirow}

\usepackage{textcase}
\usepackage[table,xcdraw]{xcolor}

% ...

\definecolor{mygray}{gray}{.9}

\ifarxiv  \fi

\newcommand{\pub}[1]{{\color{gray}{\tiny{[{#1}]\!}}}}

\newcommand{\R}[1]{{%
    \textbf{%
        \ifstrequal{#1}{1}{\textcolor{red}{R#1}}{%
        \ifstrequal{#1}{2}{\textcolor{blue}{R#1}}{%
        \ifstrequal{#1}{3}{\textcolor{magenta}{R#1}}{%
        \ifstrequal{#1}{4}{\textcolor{teal}{R#1}}{%
                           \textcolor{cyan}{R#1}%
        }}}}%
    }%
}}

\DeclareMathOperator*{\argminA}{arg\,min} 

%% file: 00_abstract.tex
\begin{abstract}

\renewcommand{\thefootnote}{\fnsymbol{footnote}}
\footnotetext[1]{Corresponding author.}
In this work, we introduce \textsc{ProMotion}, a unified prototypical transformer-based framework engineered to  model fundamental motion tasks. \textsc{ProMotion} offers a range of compelling attributes that set it apart from current task-specific paradigms. \ding{182} We adopt a prototypical perspective, establishing a unified paradigm that harmonizes disparate motion learning approaches. This novel paradigm streamlines the architectural design, enabling the simultaneous assimilation of diverse motion information. \ding{183} We capitalize on a dual mechanism involving the \emph{feature denoiser} and the \emph{prototypical learner} to decipher the intricacies of motion. This approach effectively circumvents the pitfalls of ambiguity in pixel-wise feature matching, significantly bolstering the robustness of motion representation. \ding{184} We demonstrate a profound degree of transferability across distinct motion patterns. This inherent versatility reverberates robustly across a comprehensive spectrum of both 2D and 3D downstream tasks. Empirical results demonstrate that \textsc{ProMotion} outperforms various well-known specialized architectures, achieving 0.54 and 0.054 Abs Rel error on the Sintel and KITTI depth datasets, 1.04 and 2.01 average endpoint error on the clean and final pass of Sintel flow benchmark, and 4.30 F1-all error on the KITTI flow benchmark. For its efficacy, we hope our work can catalyze a paradigm shift in universal models in computer vision.
\end{abstract}

%% file: 01_intro.tex

\section{Introduction}
\label{sec:intro}

\begin{figure*}
\centering
\includegraphics[width=0.99\linewidth, height=7.7cm]{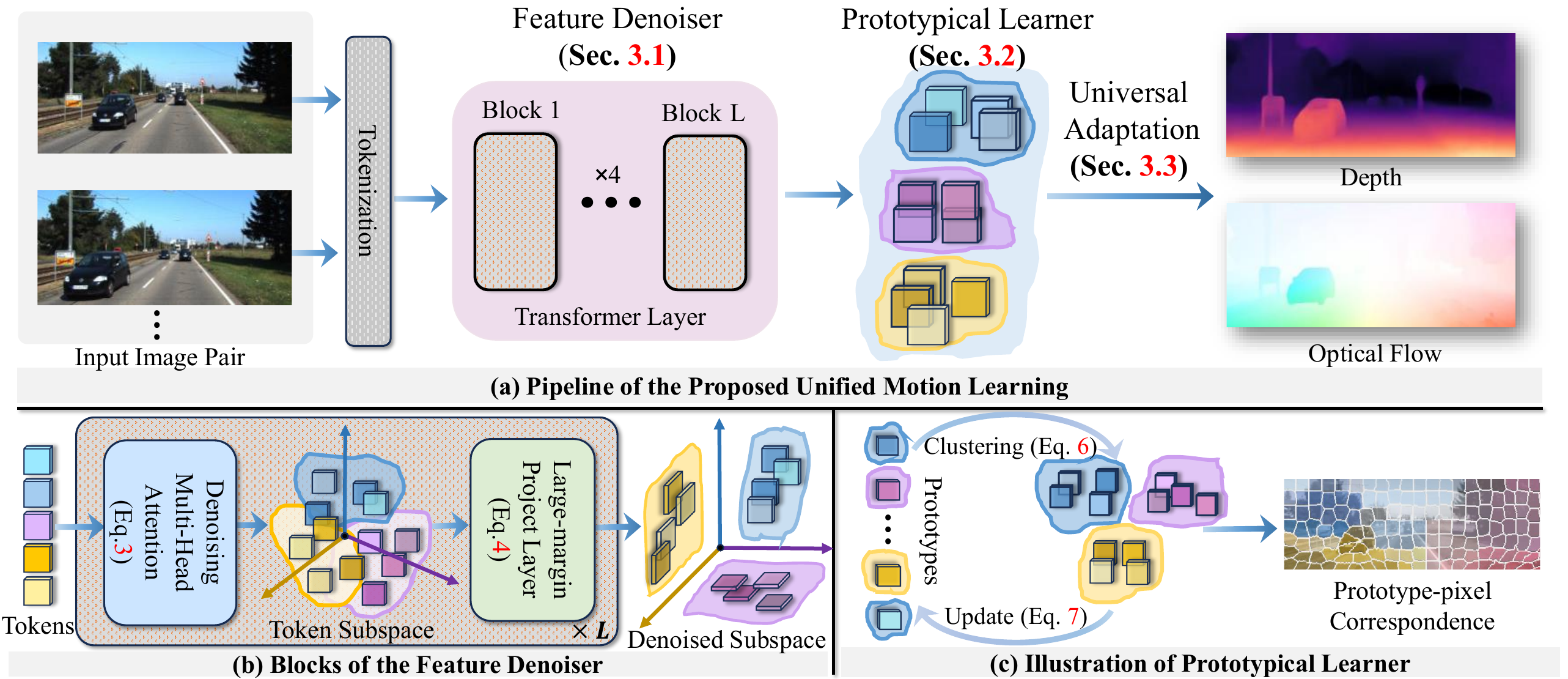}
\vspace{-1.5mm}
\caption{(a) Overall pipeline of \textsc{ProMotion}. (b) Each Transformer block in \textit{feature denoiser} maps the input tokens into different feature subspaces (Eq. \ref{denoising}) and then projects them to the orthogonal direction (Eq. \ref{projection}), therby mitigating the uncertainties in motion for robustness. (c) The \textit{prototypical learner} clusters the subspace into prototypes (Eq. \ref{cluster}) and performs iterations to update them (Eq. \ref{update}). The learned prototype can capture different motion patterns, enabling representation learning for various dynamic characteristics.
\label{fig:overview}}
\vspace{-0.1in}
\end{figure*}
\begin{verse}
    \emph{``The eventual goal of science is to provide a single theory that describes the whole universe.''}

    \hfill $-$ Stephen Hawking ~\cite{hawking1988brief}
\end{verse}

The quote coined by the famous physicist Stephen Hawking reflects that, the nature of science is to quest a \emph{unified} approach to describe and understand various phenomena in the universe. So, \emph{what about motion?}

Motion learning, such as \emph{optical flow} and \emph{scene depth}, in videos is one of the most fundamental problems in computer vision, and has a wide spectrum of key applications in many downstream tasks, \eg, 2D video segmentation and tracking and 3D detection. Current works, somehow unexpected, are often focused on developing a specialized architecture for a particular aspect in motion learning, \eg, either optical flow \cite{huang2022flowformer,shi2023flowformer++,dong2023rethinking} or scene depth~\cite{bhat2021adabins,li2023depthformer}, while neglecting the underlying connections between different motion learners and their unification for motion representation, which potentially could be transformative for motion learning. In addition, existing task-specific motion learning mechanism may lack the flexibility to generalize to broader downstream vision tasks, \eg, being applicable to either 2D or 3D tasks, \emph{not} both. Therefore, the following question, as in the above quote but for motion learning, naturally arises: \emph{\ding{172} Is there a manner to unify different motion learners?}

Besides, for motion learning tasks, the notorious photometric inconsistency (\eg, caused by the shadows and illumination variations) widely exists, which introduces unexpected uncertainty during matching and hence compromise the underlying motion representation learning by inaccurate or even false pixel-wise feature matching, degrading performance. Thus, along with question \ding{172}, another question needs to be answered for unified motion learning: \emph{\ding{173} How can we mitigate uncertainty in motion learning?} or in other words, \emph{how to learn uncertainty reduced motion representation?}

In an effort to embrace these challenges, we propose \textbf{\emph{ProMotion}}, a novel \textbf{Pro}totypical \textbf{Motion} architecture based on Transformer for unified learning of the optical flow and depth from videos. With the learned knowledge, $\emph{ProMotion}$ can seamlessly adapt to downstream tasks with ease (Fig.~\ref{fig:intro_fig}).
The key motivation behind \textsc{ProMotion} is that, motion (\eg, optical flow or scene depth) essentially is to describe movement of an object or a semantic region, which can be seen as a \emph{prototypical learner} to a set of pixel exemplars. Viewed in this light, prototype, as a underlying connection between different motion tasks, is leveraged for unified motion learning, answering the question \ding{172}. Besides, by modeling motion using semantic prototypical features instead of pixel features, and optimizing them in the designed \emph{feature denoiser}, we reduce the risk of noisy as well as outlier pixels for feature matching, which effectively alleviates the uncertainty issue, answering the question \ding{173}.

At the core of \textsc{ProMotion} lies a novel prototypical Transformer architecture. 
Different from existing task-specific motion learning models \cite{huang2022flowformer, shi2023flowformer++, li2023depthformer} that directly extract feature embeddings from Transformer, \textsc{ProMotion} gradually denoises the feature space to preserve the most dominant features and then optimizes the denoised features to subspaces in appearance, texture, and geometry along orthogonal direction, which enforces intra-class compactness. Afterwards, the optimized subspaces are grouped into a set of semantic prototypes with iterative clustering. In comparison with the feature embeddings from vanilla Transformer \cite{wang2022pvt}, the optimized subspace features from \textsc{ProMotion} are denoised to some extent and thus are more suitable for prototype learning. 
Eventually, the prototypes are fed to universal adaptation head to produce motion information such as optical flow and depth. Fig.~\ref{fig:overview} illustrates the proposed \textsc{ProMotion} for unified motion learning.

To the best of our knowledge, \textsc{ProMotion} is the first unified motion learners with prototype learning. In comparison to current task-specific designs, \textsc{ProMotion} enjoys a few attractive qualities: \ding{182} \textbf{\emph{Unified motion learning}}. \textsc{ProMotion}, from the prototype perspective, provides a single paradigm (see Fig.~\ref{fig:overview} (a)) for different motion learners, which simplifies the architecture design of motion learning and allows unified learning of various motion information. \ding{183} \textbf{\emph{Uncertainty-aware motion representation}}. By learning motion from \emph{feature denoiser} (see Fig.~\ref{fig:overview} (b)) and \emph{prototypical learner} (see Fig.~\ref{fig:overview} (c)), \textsc{ProMotion} avoids noisy and outlier pixel-wise feature matching, which mitigates the uncertainties in motion for robustness. \ding{184} \textbf{\emph{Generalization to more downstream tasks}}. \textsc{ProMotion} learns different motion information within one single framework, which could benefit more 2D and 3D downstream tasks.


To effectively assess our method, we present compelling experimental results on various datasets and different settings. We show experimentally in \S\ref{subsec:exp_flow}, with the task of optical flow estimation, \textit{ProMotion} outperforms existing counterparts, \eg, 0.16 and 0.29 lower Average End-Point Error (\textit{AEPE}) compared to Flowformer~\cite{huang2022flowformer} and MatchFlow~\cite{dong2023rethinking} on the clean pass of Sintel.
In \S\ref{subsec:exp_depth}, under the same paradigm, our depth estimation outperforms competitors by significant margins in the $Abs\ Rel$ and $Sq\ Rel$ metrics. For instance, we reduce the relative loss by 0.08 for $Abs\ Rel$ and 0.27 for $Sq\ Rel$ compared to recent DPT \cite{ranftl2021vision} on Sintel.   
Our algorithms are extensively tested, and the efficacy of the core components is demonstrated through ablative studies in \S\ref{subsec:exp_ablation}.
We believe this work can provide insights into this field.


%% file: 02_related.tex
\section{Related Work}
\label{sec:related}

\quad \textbf{Motion Task.}
Motion tasks are characterized by the intricate processes of identifying, modeling, and predicting the motion patterns of objects and scenes, making it indispensable for a wide range of computer vision applications and subsequent tasks. Motion tasks play a pivotal role in detecting the motion of vehicles and pedestrians~\cite{hu2019joint, liu2020video, xu2022vehicle, khalifa2020pedestrian, cui2023collaborative} for self-driving car navigation and perception, identifying abnormal activities in video surveillance, and recognizing actions~\cite{tudor2017unmasking, zhou2019anomalynet, li2021variational}. In addition, motion tasks improve video compression and segmentation~\cite{hu2021fvc, gao2022structure, qin2023coarse, lu2023label} by predicting object and frame-level relationships. Among the myriad of motion-related tasks, two highly representative tasks stand out as particularly noteworthy: optical flow and scene depth estimation. 

\textbf{Optical Flow and Depth.} 
Current methods for estimating optical flow and scene depth have created specialized architectures for each, ignoring their intrinsic relationships.

Optical flow, which involves finding 2D pixel displacements between images, is a fundamental problem in computer vision.
FlowNet~\cite{ilg2017flownet}, as a pioneering work, first introduced the promise of a CNN-based architecture for direct flow regression. Building upon synthetic data
for pretraining, it achieves coarse estimation performance. This sparked further explorations in architectures and training strategies, including iterative refinement and pyramid regression~\cite{sun2018pwc, ranjan2017optical, teed2020raft}, incorporating photometric~\cite{liu2019selflow} and forward-backward consistency~\cite{jonschkowski2020matters}, and transformer architecture~\cite{sui2022craft, huang2022flowformer, lu2023transflow, shi2023flowformer++}. These advances have outperformed earlier approaches by significantly reducing end-point errors.

Learning-based scene depth prediction typically takes a single image as input and utilizes generic network architectures such as ResNet~\cite{he2016deep} and ViT~\cite{dosovitskiy2020vit} to predict scene depth. DepthNet~\cite{eigen2014depth} pioneered the direct depth regression networks and established optimization through the Scale-Invariant Log Loss (SIlog). Significant progress has been driven by innovations in network architectures~\cite{laina2016deeper,xu2018structured, miangoleh2021boosting,lee2018single}, optimization schemes~\cite{laina2016deeper,fu2018deep,cheng2022physical}, and multi-view fusion~\cite{guizilini2022multi,rich20213dvnet}. To improve the architectures, Laina et al.~\cite{laina2016deeper} developed a residual network, Xu et al.~\cite{xu2018structured} presented attention-guided neural fields, and Lee et al.~\cite{lee2018single} used a multi-scale fusion scheme to improve contextual reasoning. For optimization, advanced loss functions, including reverse Huber loss~\cite{laina2016deeper} and ordinal regression~\cite{fu2018deep} provide stronger supervision signals. Multi-view feature fusion incorporates more geometric cues from additional perspectives~\cite{rich20213dvnet}.

However, unlike the specialized architectures that focus on improvements in architecture, optimization, etc., our worldview cognition is more ambitious in unifying motion tasks under one paradigm, which is a rare exploration.

\textbf{Unified Vision Models.}
As a recent trend in scientific research, unified theories have seen their emergence. There have been successful attempts to train different language modeling objectives within a single model~\cite{dong2019unified, khashabi2020unifiedqa, xie2022unifiedskg, lourie2021unicorn, khashabi2022unifiedqa}. However, the development of unified models for computer vision still lags far behind language models.
In the vision regime, some of the early research efforts have focused on the development of either encoders~\cite{dosovitskiy2020vit, liu2021swin, dong2022cswin, dehghani2023scaling} or decoders ~\cite{cheng2021mask2former, yu2022cmt, liang2022expediting, li2023mask}. The encoders focus on
the effort to develop generic backbones that are trained on extensive data, so they can be adapted for different downstream tasks. For instance, Swin~\cite{liu2021swin} and CSWin Transformer~\cite{dong2022cswin} serve as a general-purpose backbone for visual tasks.
In contrast, studies on decoders ~\cite{cheng2021mask2former, yu2022cmt, liang2022expediting, li2023mask} are designed to address homogeneous specific tasks (\eg, object recognition ~\cite{li2023mask, zhou2022simple}, instance segmentation ~\cite{cheng2021mask2former, li2023mask}, semantic segmentation ~\cite{cheng2021mask2former, li2023mask}) by representing visual patterns from queries. Despite this, there has been little effort to create a unified model for motion tasks involving optical flow and scene depth estimation.

\textbf{Prototype Learning.}
Prototype-based methods have gained substantial attention in the field of machine learning due to their alignment with human cognitive processes and intuitive appeal. Studies in psychology have demonstrated that humans tend to rely on prototypical examples as a basis for learning and problem-solving when confronted with novel challenges~\cite{newell1972human, yang2021multiple}. In the realm of machine learning, prototype-based methodologies deviate from conventional techniques, such as support vector machines~\cite{hastie2009elements} and multilayer perceptrons~\cite{bishop2006pattern}. They instead reason about observations by making comparisons with representative examples. Among the earliest prototype-based classifiers, the nearest neighbor algorithm~\cite{cover1967nearest} stands out and laid the foundation for subsequent methods like learning vector quantization (LVQ)~\cite{kohonen1990self}, generalized LVQ~\cite{sato1995generalized} and deep nearest centroids~\cite{wang2023visual}. Recent times have witnessed a surge of interest in deep learning techniques, combined with prototype learning, which have shown promising results across various domains within machine learning. These include few-shot learning~\cite{snell2017prototypical, liu2022intermediate}, zero-shot learning~\cite{jetley2015prototypical, zhang2021prototypical}, unsupervised learning~\cite{wu2018unsupervised, xu2020attribute}, and supervised learning~\cite{wu2018improving, mettes2019hyperspherical, yang2018robust}. Drawing from the valuable insights offered by these prior works, we embark on an endeavor to extend prototype learning into the motion tasks. In this context, we interpret the motion of objects or regions as an ensemble of prototypes. This innovative approach inherently encapsulates the dynamic features of motion, resulting in an advancement in the foundational learning of diverse motion attributes.

%% file: 03_method.tex
\section{ProMotion}
\label{sec:method}

In this section, we present \textsc{ProMotion}, a novel framework for unified motion tasks (see Fig.~\ref{fig:overview} (a)). The model has a serial of hierarchical transformer blocks in \textit{feature denoiser} that enables compact feature learning, followed by \textit{prototypical learner} to capture valuable motion concepts. We elaborate the pipeline in two stages --- first \textit{feature denoiser} (\S\ref{subsec:ProTer}) and then \textit{prototypical learner} (\S\ref{subsec:Prototypical}).

\subsection{Feature Denoiser}
\label{subsec:ProTer}

\textbf{Proposition 1.} \textit{Feature denoising enhances representation learning by capturing more informative subspaces.}

To suppress noises and mitigate uncertainty in motion representation learning, we propose a novel \emph{feature denoiser} for enhancement. 
Specifically, considering a mapping function $\Phi(\cdot)$ and input patch tokens $X \in \mathbb{R}^{D \times N}$, the goal of feature denoiser is to learn a more compact and dominant feature representation $Z \in \mathbb{R}^{d \times N}$ via \emph{subspace denoising} and \emph{large-margin projection}:
\begin{equation}
\vspace{-1mm}
\Phi: x \in \mathbb{R}^{D \times N}  \rightarrow   z \in \mathbb{R}^{d \times N}, \quad d<D
\end{equation}

For subspace denoising, we abstract the raw representation $x$ using the expected denoised tokens $z$ with an additive Gaussian noise $w$ as: $x = z + \sigma  w$, where $w \sim N(0, 1)$. 
The objective is to minimize the distance between raw representation $x$ and clean representation $z$, under noise level 
$\sigma^{l}$ as:
\begin{equation}\label{opt}
E[z | \cdot] = \argminA_f E_{z,w} [\left \| f(z+\sigma^{l} w)  - z \right \|_{2}^{2}]
\end{equation}
where $f(\cdot)$ is the transition function at the $l^{\text{th}}$ transformer layer. If $z^{l}$ is the feature representation under noise level $\sigma^{l}$, and the updated representation at the next stage is $z^{l'} = E[z|z^{l}]$. Using Tweedie's formula~\cite{efron2011tweedie}, the optimization in Eq.~\ref{opt} can be solved by computing posterior expectations as:
\begin{equation}
\vspace{-1mm}
\begin{split}
z^{l'} & = z^{l} + (\sigma^{l})^2  \nabla_{\boldsymbol{x}} \log p\left(z^l\right) \\
& \approx  \sum_{k=1}^{K} (w_k * U_k * (U_k^T * z^l))
\end{split}
\label{denoising}
\end{equation}
where $w_k$ satisfies 
$Softmax(\frac{1}{2 \times (\sigma^{l})^{2}} * ||U_{k}^{T} z^{l}||_2^{2})$. 
This reveals that the signal $z^{l'}$ can be naturally reconstructed as a weighted sum of multiple linear projections $U_k * (U_k^T * z^l)$, which enables reformulation of the transformer encoder to avoid non-linear attention blocks, much simpler yet effective.

To improve the robustness of feature embeddings to noise, we further apply a large-margin projection to enforce inter-class separability and intra-class clustering that maps features in orthogonal directions. For this purpose,
the Iterative Shrinkage Thresholding Algorithm (ISTA) \cite{beck2009fast, gregor2010learning, wright2022high, yu2023white} is used to infer the $z^{l+1}$ as:
\begin{equation}
\vspace{-1mm}
\begin{split}
z^{l+1} &= ISTA(z^{l'} | \ \boldsymbol{O}^{l}) \\
        &= \mathrm{ReLU}\left(z^{l'}+\epsilon \boldsymbol{O}^{l*}\left(z^{l'}-\boldsymbol{O}^{l} z^{l'}\right)\right)
\end{split}   
\label{projection}
\end{equation}
where $\epsilon$ $>$ 0 is a step size and $O^{l}$ is the sparsifying orthogonal dictionary. With this approach, the projected subspace can be interpreted as ReLU activation function of multiple linear transformations.
The design of subspace denoising and large-margin projection aligns with the trend of achieving representative representation through sparsifying representation, which enjoys the following appealing characteristics:

\begin{itemize}
    \item \textit{Underlying Data Structure:} Subspace denoising
    maps the input tokens to different feature subspaces and removes irrelevant variances, allowing the most representative data structure to emerge clearly. The large-margin projection further projects feature subspaces in a low-dimensional orthogonal direction to increase inter-class separation. As a result, \textit{Feature Denoiser} reflects the most representative or characteristic relationships.
        \item \textit{Uncertainty Mitigation:} Aligning closely with the underlying data structure, subspace denoising improves the signal-to-noise ratio and allows for more repeatable matching correspondences, while large-margin projection eliminates redundant dimensions, allowing for more compact features for matching. \textit{Feature Denoiser} mitigates uncertainty after subspace denoising and large-margin projection.
\end{itemize}

\subsection{Prototypical Learner}
\label{subsec:Prototypical}
\textbf{Proposition 2.} 
\textit{Dynamic patterns can be captured by a set of prototypes, therefore unifying different motion tasks.}

After subspace denoising in the \textit{feature denoiser}, we propose to employ prototypical learning on the denoised and compact representation for unified motion learning.
The essence of prototypical learning is that, for every token embedding $\mathcal{X}$, find a mapping function $f_{\mathrm{c}}$ to generate $K$ prototypes with prototype centers $C \in \mathbb{R}^{K\times D}$ that is initialized from a $D$-dimensional image
grid. For each image $x_{i}$, the expected set of prototype segments are: 
\begin{equation}
\vspace{-1mm}
   f_{\mathrm{c}}(\mathcal{X})=\left(\boldsymbol{r}_i\right)_{i=1}^N , 
 \qquad \boldsymbol{r}_i=\left(r_{i, k}\right)_{k=1}^K
 \label{initilization}
\end{equation}

The initial prototypes are iteratively updated by Sinkhorn-Knopp based clustering
as an approximate solution of optimal transport problem:
\begin{equation}
\vspace{-1mm}
\max _{\mathbf{Q} \in Q} \operatorname{Tr}\left(\mathbf{Q}^{\top} \mathbf{V}\right), 
\quad s.t. \ \mathbf{Q} \mathbf{1}_N = \frac{1}{K} \mathbf{1}_{K}, \mathbf{Q}^{\top} \mathbf{1}_{K}=\frac{1}{N} \mathbf{1}_{N}
\label{cluster}
\end{equation}
where $\mathbf{Q} \in \mathbb{R}^{K \times N}$ is the soft cluster assignment matrix and is constrained in the transportation polytope under the equipartition constraint~\cite{distances2013lightspeed}, $\mathbf{V}$ is the similarity matrix where $\mathbf{V} = \mathbf{C} \mathcal{X}^{T} \in \mathbb{R}^{K \times N}$. 
Current prototypes are updated according to  $\mathbf{Q}_{t-1}$ for generating new prototypes centered at $C_{t}$:
\begin{equation}
C_{t} \leftarrow  \mathbf{Q}_{t-1} \mathcal{X} \  \in  \ \mathbb{R}^{K \times D}  
\label{update}
\end{equation}

By modeling the intrinsic similarities of prototypes via prototypical learning, the method refines the feature embeddings and improves the overall understanding of the underlying context and structure. All feature representations can be seamlessly used for various target tasks via adaptation heads (see \S\ref{subsec:implementation_details}). Prototypical learning has the following features:

\begin{itemize}
   \item \textit{Unified Paradigm:} 
   Prototypical learning promotes the unified paradigm to encapsulate the underlying structure and context of data that can be leveraged across a variety of data and visual tasks. The learned prototypes capture the most representative and informative prototype-wise structures that are invariant across diverse data distributions and tasks, leading to great generalizability and adaptability.

        \item \textit{Transferability:} 
           The prototypes generated capture the underlying data structures and the essence of a category, which can be used directly by adding task-specific heads for downstream tasks. For complex data or tasks, the learned prototypes can also be reused by adding new prototypes or re-weighting the existing ones. This elegant design facilitates representation learning for knowledge transferability in accordance with the unified paradigm.

\end{itemize}

\subsection{Implementation Details}
\label{subsec:implementation_details}
The implementation details and overall framework of \textsc{ProMotion} are shown in Fig. 
\ref{fig:overview}.
\begin{itemize}
    \item \textit{Encoder.}
    The goal of the encoding process is to generate token prototypes from the given image $I$. The whole pipeline starts with tokenization to convert the $I$ into token embeddings. Subsequently, the tokens are passed to \textit{feature denoiser} (subspace denoising (Eq. \ref{denoising}) and large-margin projection (Eq. \ref{projection})) to learn denoised and  compact subspaces. Once the denoised subspaces are learned, a set of prototype features is inferred from the subspaces (Eq. \ref{initilization}), and are then updated via iterative clustering (Eq. \ref{update}).

        \item \textit{Universal Adaptation.} For optical flow and scene depth,
            the flow and depth heads upsample the features of the token prototypes by predicting a convex mask to retrieve the flow map and depth. The initial predictions $(\hat{f}_0, \hat{d}_0)$ are updated by continuously estimating the residuals $(\delta \hat{f}, \delta \hat{d})$ \cite{teed2020raft, zhou2021r}. The last estimate is used as the final prediction as $\hat{f}_n, \hat{d}_n = \Delta \hat{f}, \Delta \hat{d} + \hat{f}_{n-1}, \hat{d}_{n-1}$.             
            When transferring to downstream tasks, the pre-trained motion features are frozen, and only the task-specific adaptation head is fine-tuned,            
leveraging the core designs of the \textit{feature denoiser} (\S\ref{subsec:ProTer}) and \textit{prototypical learner} (\S\ref{subsec:Prototypical}).

               \item \textit{Loss Functions.} We use the SILog loss~\cite{fu2018deep} and a weighted $L_{1}$ loss \cite{teed2020raft} for guiding the depth and flow learning. 

\end{itemize}

\begin{table*}[tb] \small
\centering
\scalebox{0.92}{\begin{tabular}{clccccccc}
\hline
\rowcolor{mygray}
\multirow{2}{*}{Training} & \multirow{2}{*}{Method} & \multicolumn{2}{c}{\uline{Sintel (val)}} & \multicolumn{2}{c}{\uline{KITTI-15 (val)}} & \multicolumn{2}{c}{\uline{Sintel (test)}} & \uline{KITTI-15 (test)}\tabularnewline
\rowcolor{mygray}
 \multirow{-2}{*}{Training}& \multirow{-2}{*}{Method} & Clean & Final & Fl-epe & Fl-all & Clean & Final & Fl-all\tabularnewline
\hline  \hline
\multirow{15}{*}{C+T} & PWC-Net\pub{CVPR2018}~\cite{sun2018pwc} & 2.55 & 3.93 & 10.35 & 33.7 & - & - & -\tabularnewline
 & LiteFlowNet\pub{TPAMI2020}~\cite{hui2018liteflownet} & 2.24 & 3.78 & 8.97 & 25.9 & - & - & -\tabularnewline
 & RAFT\pub{ECCV2020}~\cite{teed2020raft} & 1.43 & 2.71 & 5.04 & 17.4 & - & - & -\tabularnewline
 & Separable Flow\pub{ICCV2021}~\cite{zhang2021separable} & 1.30 & 2.59 & 4.60 & 15.9 & - & - & -\tabularnewline
 & GMA\pub{ICCV2021}~\cite{jiang2021learning} & 1.30 & 2.74 & 4.69 & 17.1 & - & - & -\tabularnewline
 & AGFlow\pub{AAAI2022}~\cite{luo2022learning} & 1.31 & 2.69 & 4.82 & 17.0 & - & - & -\tabularnewline
 & KPA-Flow\pub{CVPR2022}~\cite{luo2022kpa} & 1.28 & 2.68 & 4.46 & 15.9 & - & - & -\tabularnewline
 & DIP\pub{CVPR2022}~\cite{zheng2022dip} & 1.30 & 2.82 & 4.29 & \uline{13.7} & - & - & -\tabularnewline
 & GMFlowNet\pub{CVPR2022}~\cite{zhao2022global} & 1.14 & 2.71 & 4.24 & 15.4 & - & - & -\tabularnewline
 & GMFlow\pub{CVPR2022}~\cite{xu2022gmflow} & 1.08 & 2.48 & 7.77 & 23.40 & - & - & -\tabularnewline
 & CRAFT\pub{CVPR2022}~\cite{sui2022craft} & 1.27 & 2.79 & 4.88 & 17.5 & - & - & -\tabularnewline
 & FlowFormer\pub{ECCV2022}~\cite{huang2022flowformer} & \uline{1.01} & \uline{2.40} & \uline{4.09}$^{\dagger}$ & {14.7}$^{\dagger}$ & - & - & -\tabularnewline
 & SKFlow\pub{NIPS2022}~\cite{sun2022skflow} & 1.22 & 2.46 & 4.27 & 15.5 & - & - & -\tabularnewline
 & MatchFlow\pub{CVPR2023}~\cite{dong2023rethinking}& 1.14 & 2.61 & 4.19$^{\dagger}$ & \textbf{13.6}$^{\dagger}$ & - & - & -\tabularnewline
  & \textbf{Ours} & \textbf{0.91} & \textbf{2.32} & \textbf{3.95} & 14.1 & - & - & -\tabularnewline

\hline
\multirow{15}{*}{C+T+S+K(+H)} 
 & PWC-Net\pub{CVPR2018}~\cite{sun2018pwc} & - & - & - & - & 4.39 & 5.04 & 9.60 \tabularnewline
 & RAFT\pub{ECCV2020}~\cite{teed2020raft} & (0.76) & (1.22) & (0.63) & (1.5) & 1.61 & 2.86 & 5.10\tabularnewline
 & Separable Flow\pub{ICCV2021}~\cite{zhang2021separable} & (0.69) & (1.10) & (0.69) & (1.60) & 1.50 & 2.67 & {4.64}\tabularnewline
 & GMA\pub{ICCV2021}~\cite{jiang2021learning} & (0.62) & (1.06) & (0.57) & (1.2) & 1.39 & 2.47 & 5.15\tabularnewline
 & AGFlow\pub{AAAI2022}~\cite{luo2022learning} & (0.65) & (1.07) & (0.58) & (1.2) & 1.43 & 2.47 & 4.89\tabularnewline
 & KPA-Flow\pub{CVPR2022}~\cite{luo2022kpa} & (0.60) & (1.02) & (0.52) & (1.1) & 1.35 & 2.36 & 4.60\tabularnewline
 & DIP\pub{CVPR2022}~\cite{zheng2022dip} & - & - & - & - & 1.44 & 2.83 & \textbf{4.21}\tabularnewline
 & GMFlowNet\pub{CVPR2022}~\cite{zhao2022global} & (0.59) & (0.91) & (0.64) & (1.51) & 1.39 & 2.65 & 4.79\tabularnewline
 & GMFlow\pub{CVPR2022}~\cite{xu2022gmflow} & - & - & - & - & 1.74 & 2.90 & 9.32\tabularnewline
 & CRAFT\pub{CVPR2022}~\cite{sui2022craft} & (0.60) & (1.06) & (0.58) & (1.34) & 1.45 & 2.42& 4.79\tabularnewline
 & FlowFormer\pub{ECCV2022}~\cite{huang2022flowformer} & (\uline{0.48}) & (\uline{0.74}) & (0.53) & (1.11) & \uline{1.20} & \uline{2.12} & 4.68$^{\dagger}$\tabularnewline
 & SKFlow\pub{NIPS2022}~\cite{sun2022skflow} & (0.52) & (0.78) & (\uline{0.51}) & (\textbf{0.94}) & 1.28& 2.23& 4.84\tabularnewline
 & MatchFlow\pub{CVPR2023}~\cite{dong2023rethinking} & (0.51) & (0.81) & (0.59) & (1.3) & 1.33 & 2.64& 4.72 \tabularnewline
& \textbf{Ours} & (\textbf{0.40}) & \textbf{0.66} & \textbf{0.48} & \uline{0.98} & \textbf{1.04} & \textbf{2.01} & \uline{4.30} \tabularnewline
\bottomrule
\end{tabular}}
\vspace{-0.1in}
\caption{\textbf{Quantitative comparison on Sintel.} `C+T': Succeeding training on FlyingChairs (C) and FlyingThings3D (T), the models are evaluated for generalization on Sintel (S) and KITTI (K) validation sets. `C+T+S+K(+H)': Training on a combination of C, T, S, K, and HD1K (+H) is evaluated. The first and second place results are \textbf{bolded} and \uline{underlined}, respectively. $^\dagger$ indicates the use of tile technique~\cite{jaegle2021perceiver}.
\label{tab:flow_results}}
\vspace{-0.2in}
\end{table*}

\section{Experiments}
\label{sec:exp}

\subsection{Experiments on Optical Flow}
\label{subsec:exp_flow}


\textbf{Training.} We follow standard optical flow training steps~\cite{teed2020raft, huang2022flowformer} to train our model on FlyingChair (C) for 80K iterations and FlyingThings3D (T) for another 80K iterations (denoted as `C+T') with batch size of 16. Following this, we fine-tune the model on a larger combination set of FlyingThings3D (T), Sintel (S), KITTI (K) and HD1K (H) (`C+T+S+K+H'). The method is implemented on 8 NVIDIA A100 GPUs with one-cycle learning strategy~\cite{smith2019super} and AdamW optimizer~\cite{loshchilov2017decoupled} with a maximum learning rate of 3$e$-4.

\noindent \textbf{Testing.} This model is evaluated on Sintel and KITTI benchmarks. Generalization and ablation are on the validation split of Sintel and KITTI following~\cite{teed2020raft, huang2022flowformer}.

\noindent \textbf{Metrics.}
Following the evaluation metrics from the Sintel and KITTI benchmarks, we use the average end-point error (\textit{AEPE}) to measure the average pixel-wise flow error on Sintel, and \textit{Fl-epe} (\textit{AEPE} on outlier pixels) and \textit{Fl-all} (\textit{AEPE} over all pixels) on KITTI.  

\noindent \textbf{Quantitative Results.}
Following~\cite{teed2020raft}, we first evaluate the generalization performance after training on C+T. As shown in Table~\ref{tab:flow_results}, \textsc{ProMotion} achieves a 36.4\% and 20.2\% reduction in \textit{AEPE} on Sintel clean pass, surpassing both the strong baseline \cite{teed2020raft} and the recent SOTA method \cite{dong2023rethinking}. On KITTI, there is a 21.6\% and 5.7\% decrease of \textit{Fl-epe} in comparison to \cite{teed2020raft} and \cite{dong2023rethinking}. These outcomes indicate the good generalization of our approach.

The method is then evaluated in a more general manner to report leaderboard results for the Sintel and KITTI test sets with training on C+T+S+K(+H). On the Sintel, \textsc{ProMotion} obtains \textit{AEPE} of 1.04 and 2.01 on Clean and Final passes respectively, which significantly lowers baseline model~\cite{teed2020raft} error by 35.4\% and 29.7\%. Similarly, for the KITTI, \textsc{ProMotion} reduces~\cite{teed2020raft} by 15.7\%, from 5.10 to 4.30. The results demonstrate overall superior performance compared to other recent published works.

\begin{figure*}
\centering
\includegraphics[width=0.93\linewidth, height=6.7cm]{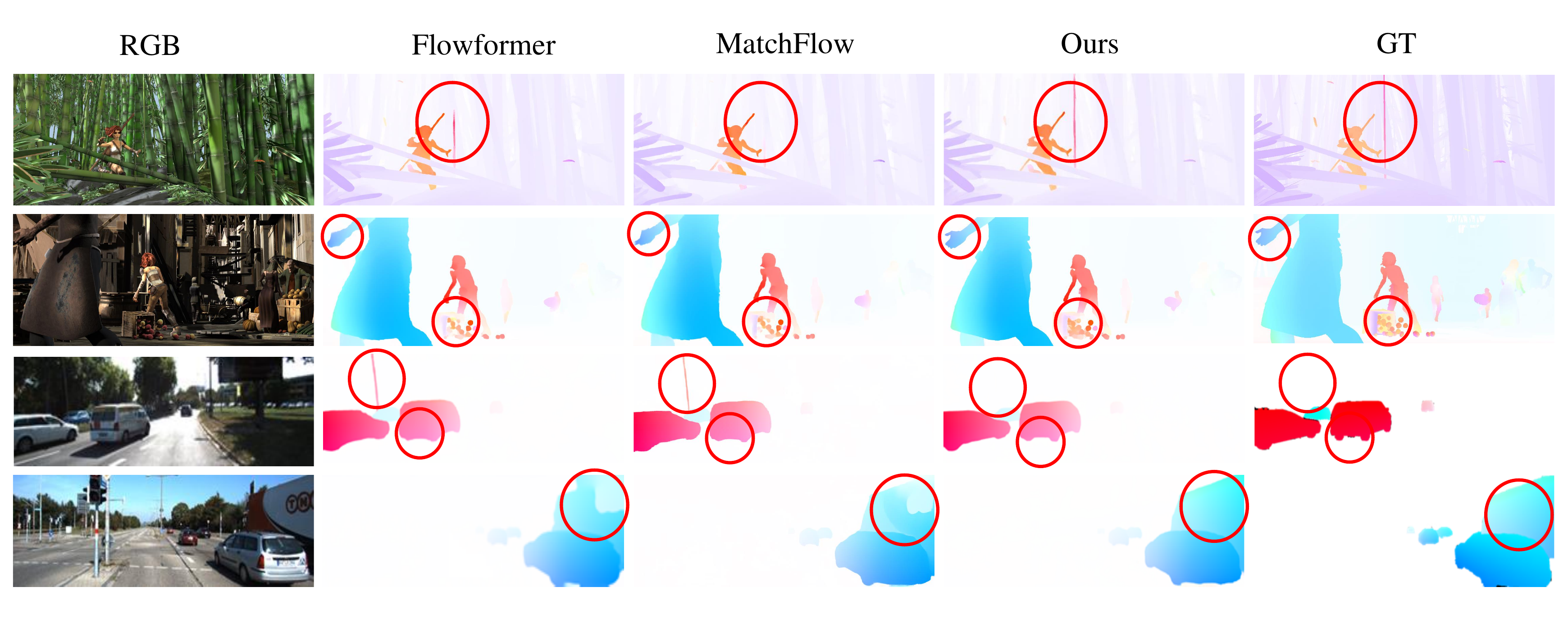}
\vspace{-0.26in}
\caption{\textbf{Qualitative comparison of optical flow on Sintel and KITTI \texttt{val} set}. Notable areas are marked with red circles. Compared to \cite{huang2022flowformer} and \cite{dong2023rethinking}, our approach shows better ability to reduce matching uncertainties due to similar patterns, illumination changes, shadows, etc. 
\label{fig:flow_qualitative}}
\vspace{-2mm}
\end{figure*}

\noindent \textbf{Qualitative Examples.}
We also provide qualitative results on Sintel and KITTI in Fig.~\ref{fig:flow_qualitative}. Under severe illumination and shadow variations and object occlusion, the estimation from \textsc{ProMotion} exhibits more complete objects and robust estimation (second and third rows), which is attributed to the design of \textit{feature denoiser} to mitigate uncertainties. Within highly similar textures and patterns (first and fourth rows), the estimation of \textsc{ProMotion} shows better capability in capturing object-level shapes and boundaries, which can be explained by the design of \textit{prototypical learner} to capture more object-level motion patterns.

\begin{table}[htb]
\centering
\resizebox{0.46\textwidth}{!}{  
\begin{tabular}{llccc}
\hline
\rowcolor{mygray}
Dataset              & Method      & \multicolumn{1}{c}{\cellcolor{mygray}Abs Rel $\downarrow$} & \multicolumn{1}{c}{\cellcolor{mygray} Sq Rel $\downarrow$} & \multicolumn{1}{c}{\cellcolor{mygray} RMSE $\downarrow$} \\ \hline  \hline
                     & TransDepth\pub{ICCV2021}~\cite{yang2021transformer}        &                               0.78                      &        1.07                                            &      0.83                                            \\
                     & AdaBins\pub{CVPR2021}~\cite{bhat2021adabins}     &                             0.73                        &      0.74                                              &    0.57                                              \\
Sintel               & DPT-Hybrid\pub{ICCV2021}~\cite{ranftl2021vision}     &                                              \uline{0.62}       &   \uline{0.65}                                                &                                            0.58      \\
                     & P3Depth\pub{CVPR2022}~\cite{P3Depth}&                                         0.65           &        0.67                                             &    \uline{0.39}                                              \\
                     & Ours        &    \textbf{0.54}                                                 &                   \textbf{0.38}                                 &                                 \textbf{0.36}                 \\ \hline
                     & TransDepth\pub{ICCV2021}~\cite{yang2021transformer}         &     0.064                                                &                  0.252                                  &     2.755                                             \\
                     & AdaBins\pub{CVPR2021}~\cite{bhat2021adabins}     &       \uline{0.058}                                              &                    0.198                                &    2.360                                              \\
KITTI                & DPT-Hybrid\pub{ICCV2021}~\cite{ranftl2021vision}    &      0.059                                               &              \uline{0.190}                                    &    \uline{2.315}                                              \\
                     & P3Depth\pub{CVPR2022}~\cite{P3Depth} &     0.060                                                &                 0.206                                   &      2.519                                           \\
                     & Ours        &       \textbf{0.054}                                              &    \textbf{0.182}                                                &   \textbf{2.298}                                               \\ 
\hline
\end{tabular}
}
\vspace{-2mm}
\caption{\textbf{Depth performance on Sintel clean and KITTI datasets}. The best results are in \textbf{bold} and the second-best results are \uline{underlined}. Our method outperforms other recent methods on most of the common error metrics.}
\vspace{-3mm}
\label{table:depth_results}
\end{table}

\subsection{Experiments on Scene Depth}
\label{subsec:exp_depth}


\textbf{Training.} We train scene depth under the same architecture as for flow, except for the dimensions in the headers. We follow the standard Eigen split~\cite{eigen2014depth} for training and then fine-tune the model to the smaller MPI Sintel~\cite{Butler:ECCV:2012} consisting of 14 sequences with clear edges and varying levels of motions.

\noindent \textbf{Testing.} The model is tested on the Sintel depth benchmark (1,104 images) and KITTI depth split (697 images).

\noindent \noindent \textbf{Metrics.}
Following standard depth estimation metrics, we use absolute mean relative error ($Abs\ Rel$), squared mean relative error ($Sq\ Rel$), and root mean square error ($RMSE$).

\noindent \textbf{Quantitative Results.}
As shown in Table~\ref{table:depth_results}, although the leading algorithms approach a saturation point to some extent ($Abs\ Rel$ and $Sq\ Rel$ are close among approaches), \textsc{ProMotion} showed improvements with reductions of 26.0\% and 16.9\% on Sintel and 6.9\% and 10.0\% on KITTI in $Abs\ Rel$ compared to the strong baseline AdaBins \cite{bhat2021adabins} and recent SOTA \cite{P3Depth} for monocular depth estimation. This suggests that our method is suitable for a range of depth scenarios, including synthetic Sintel and real-world KITTI.

\begin{figure*}
\centering
\includegraphics[width=0.93\linewidth, height=6.4cm]{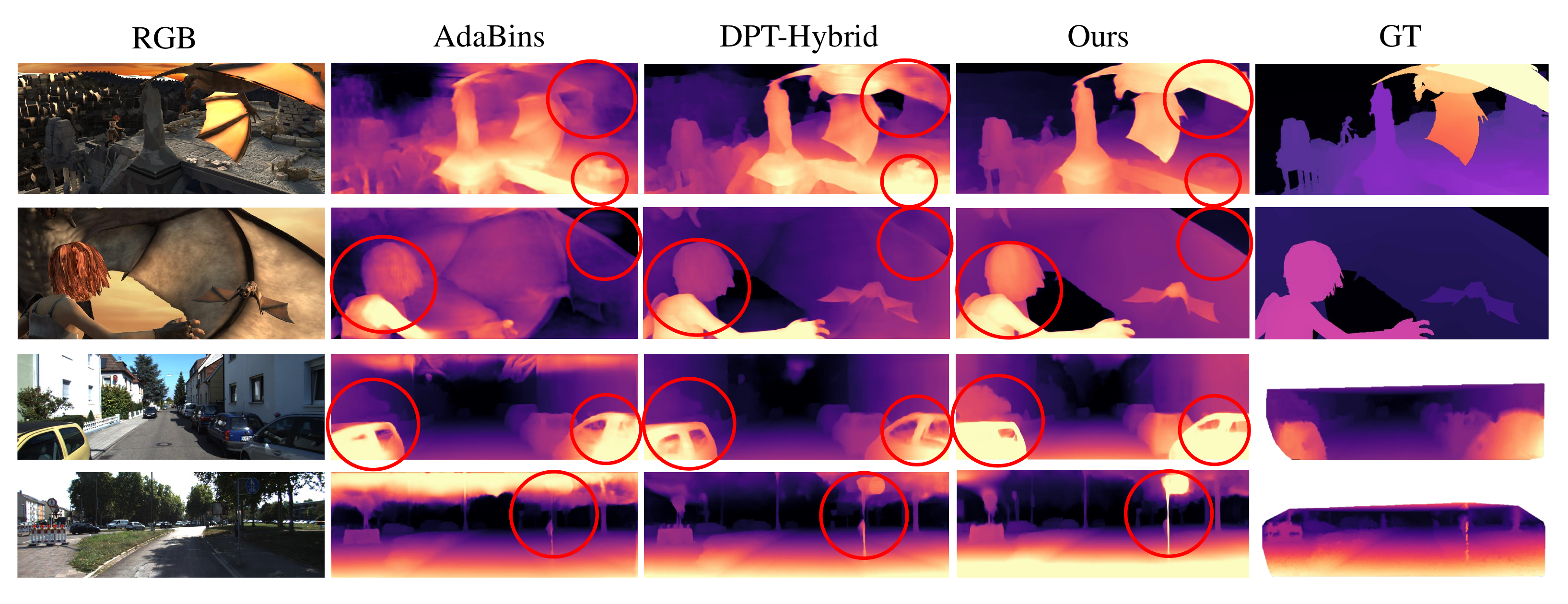}
\vspace{-0.20in}
\caption{\textbf{Qualitative comparison of scene depth on the Sintel and KITTI \texttt{val} set}. Notable areas are marked with red circles. Compared to \cite{bhat2021adabins} and \cite{ranftl2021vision}, our approach produces more consistent and smooth depths with complete object shapes and clear boundaries. Sparse ground truths in KITTI are interpolated for better visualization. 
\label{fig:depth_qualitative}}
\vspace{-1mm}
\end{figure*}

\noindent \textbf{Qualitative Examples.}
As depicted in Fig.~\ref{fig:depth_qualitative}, the compared methods AdaBins~\cite{bhat2021adabins} and DPT-Hybrid~\cite{ranftl2021vision} tend to generate inconsistent noisy depth due to the texture-less regions, shadows, and lighting changes (first, second, and third rows), while \textsc{ProMotion} shows more consistent and smoother depth estimates without being affected by the uncertainties, which is in line with the goal of our design \textit{feature denoiser}. Moreover, the proposed method is able to preserve more object characteristics (\eg, shape, surface, and boundaries) for both foreground objects and the out-of-focus image background (first and fourth rows) compared to AdaBins and DPT-Hybrid. This may be explained by the design of \textit{prototypical learner} for better object-level representation learning.

\begin{table*}[htb]	

\begin{subtable}{0.52\linewidth}   
						\captionsetup{width=.95\linewidth}
						\resizebox{\textwidth}{!}{
							\setlength\tabcolsep{4pt}
							\renewcommand\arraystretch{1.1} 
							\begin{tabular}{l|cc|cc}
\hline  
\rowcolor{mygray}   
\cellcolor{mygray}                                   & \multicolumn{2}{c|}{\cellcolor{mygray}Flow (val)}                                                & \multicolumn{2}{c}{\cellcolor{mygray}Depth (val)}                                               \\
\rowcolor{mygray}   
\multirow{-2}{*}{\cellcolor{mygray}  Algorithm Component} & \multicolumn{1}{c}{\cellcolor{mygray}  Clean} & \multicolumn{1}{c|}{\cellcolor{mygray}  Final} & \multicolumn{1}{c}{\cellcolor{mygray}  Clean} & \multicolumn{1}{c}{\cellcolor{mygray}  Final} \\ \hline \hline
\textsc{Baseline} 	
&         0.54                                          &    0.77                                               &     0.72                                              &         0.71                                          \\ 
\arrayrulecolor{gray}\hdashline\arrayrulecolor{black}
$+$ Feature Denoiser                                          &                        0.51                           &      0.73                                             &       0.67                                            &       0.66                                            \\
$+$ Prototypical Learner                                     &                     0.45                              &      0.70                                             &    0.63                                               &          0.62                                        \\
\textcolor{red}{\textbf{$+$ Unified Learning (Our ProMotion)}}   &                                 \textbf{0.40}                  &      \textbf{0.66}                                              &   \textbf{0.59}                                               &   \textbf{0.57}                                              \\ \hline 
\end{tabular}}
				\caption{\small{\textit{Key Component Analysis (\S \ref{sec:method})}}}
						\label{table:key}
\end{subtable}  
\begin{subtable}{0.45\linewidth}   
						\captionsetup{width=.95\linewidth}
						\resizebox{\textwidth}{!}{
							\setlength\tabcolsep{4.0pt}
							\renewcommand\arraystretch{1.13}
							\begin{tabular}{l|ll|ll}
\hline 
\rowcolor{mygray}  
\cellcolor{mygray}
& \multicolumn{2}{c|}{\cellcolor{mygray}Flow (val)}                                                & \multicolumn{2}{c}{\cellcolor{mygray}Depth (val)}                                               \\
\rowcolor{mygray}  
\multirow{-2}{*}{\cellcolor{mygray}Attention Types} & \multicolumn{1}{c}{\cellcolor{mygray}Clean} & \multicolumn{1}{c|}{\cellcolor{mygray}Final} & \multicolumn{1}{c}{\cellcolor{mygray}Clean} & \multicolumn{1}{c}{\cellcolor{mygray}Final} \\ \hline  \hline
Vanilla Self-Attention                                    &                          0.45                         &    0.69                                                &  0.63                                                 &            0.61                                       \\
Multi-head Self Attention                               &                           0.42                        &     0.67                                               &       0.61                                            &        0.58                                           \\
Spatial Reduction Attention                             &  0.43                                                 &  0.68                                                  & 0.63                                                  &     0.60                                              \\
\textcolor{red}{\textbf{Subspace Self-Attention (Ours)}}     &                                 \textbf{0.40}                  &      \textbf{0.66}                                              &   \textbf{0.59}                                               &   \textbf{0.57}                                                  \\ \hline
\end{tabular}}  
						\caption{\small{\textit{Attention Types (\S \ref{subsec:ProTer})}}}
						\label{table:attention}
\end{subtable}

\begin{subtable}{0.49\linewidth}    
						\captionsetup{width=.95\linewidth}
						\resizebox{\textwidth}{!}{
							\setlength\tabcolsep{5.5pt}
							\renewcommand\arraystretch{0.9}
							\begin{tabular}{l|ll|ll}
\hline
\rowcolor{mygray} 
\cellcolor{mygray}                                    & \multicolumn{2}{c|}{\cellcolor{mygray} Flow (val)}                                                & \multicolumn{2}{c}{\cellcolor{mygray} Depth (val)}                                               \\
\rowcolor{mygray} 
\multirow{-2}{*}{\cellcolor{mygray} Variant Number of Prototype} & \multicolumn{1}{c}{\cellcolor{mygray} Clean} & \multicolumn{1}{c|}{\cellcolor{mygray} Final} & \multicolumn{1}{c}{\cellcolor{mygray} Clean} & \multicolumn{1}{c}{\cellcolor{mygray} Final} \\ \hline  \hline
10                                                          &                    0.46                               &       0.71                                            &   0.65                                                &       0.62                                            \\
50                                                          &                        0.42                           &     0.69                                               &   0.61                                                &                      0.59                             \\
\textcolor{red}{\textbf{100}}                                  &                          \textbf{0.40}                         &   \textbf{0.66}                                                &     \textbf{0.59}                                              &               \textbf{0.57}                                \\
200                                &      0.41                                             &  0.68                                                  &                  0.61                                 &     0.58                                              \\ \hline
\end{tabular}}
						\caption{\small{\textit{Number of Prototypes (\S \ref{subsec:Prototypical})}}}
						\label{table:nums_proto}
\end{subtable}  \quad      
\begin{subtable}{0.46\linewidth}   
						\captionsetup{width=.95\linewidth}
						\resizebox{\textwidth}{!}{
							\setlength\tabcolsep{5.5pt}
							\renewcommand\arraystretch{0.88}
							\begin{tabular}{l|ll|ll}
\hline
\rowcolor{mygray} 
\cellcolor{mygray}                                   & \multicolumn{2}{c|}{\cellcolor{mygray}Flow (val)}                                                & \multicolumn{2}{c}{\cellcolor{mygray}Depth (val)}                                               \\  
\rowcolor{mygray} 
\multirow{-2}{*}{\cellcolor{mygray}Decoder Iteration Number} & \multicolumn{1}{c}{\cellcolor{mygray}Clean} & \multicolumn{1}{c|}{\cellcolor{mygray}Final} & \multicolumn{1}{c}{\cellcolor{mygray}Clean} & \multicolumn{1}{c}{\cellcolor{mygray}Final} \\ \hline  \hline
1                                                          &                    0.58                               &       0.89                                             &   0.84                                                &       0.77                                            \\
6                                                          &                    0.42                               &    0.71                                                &     0.62                                              &       0.61                                            \\
\textcolor{red}{\textbf{12}}                                  &                0.40                                &       \textbf{0.66}                                             &     \textbf{0.59}                                             &        0.57                                           \\
24                                 &    \textbf{0.39}                                               &    0.66                                               &     0.60                                              &                    \textbf{0.56}                               \\ \hline
\end{tabular}}
						\caption{\small{\textit{Decoder Iteration Number (\S \ref{subsec:implementation_details})}}}
						\label{table:num_iteration}
\end{subtable}    

\begin{subtable}{0.54\linewidth}   
						\captionsetup{width=.95\linewidth}
						\resizebox{\textwidth}{!}{
							\setlength\tabcolsep{5.5pt}
							\renewcommand\arraystretch{0.95}
							\begin{tabular}{l|ll|ll}
\hline
\rowcolor{mygray} 
\cellcolor[HTML]{EFEFEF}                                   & \multicolumn{2}{c|}{\cellcolor[HTML]{EFEFEF}Flow (val)}                                                & \multicolumn{2}{c}{\cellcolor[HTML]{EFEFEF}Depth (val)}                                               \\
\rowcolor{mygray} 
\multirow{-2}{*}{\cellcolor[HTML]{EFEFEF}Variant Prototype Updating Strategy} & \multicolumn{1}{c}{\cellcolor[HTML]{EFEFEF}Clean} & \multicolumn{1}{c|}{\cellcolor[HTML]{EFEFEF}Final} & \multicolumn{1}{c}{\cellcolor[HTML]{EFEFEF}Clean} & \multicolumn{1}{c}{\cellcolor[HTML]{EFEFEF}Final} \\ \hline  \hline
Cosine Similarity                                          &                          0.43                         &       0.70                                             &    0.62                                               &             0.61                                      \\
K-Means                                                    &                          0.41                         &      0.68                                              &     0.62                                              &               0.60                                    \\
Gaussian Mixture Models                                    &                    0.41                               &    0.67                                                &    0.61                                               &        0.58                                           \\
\textcolor{red}{\textbf{Sinkhorn-Knopp}}                      &                 \textbf{0.40}                                 &  \textbf{0.66}                                               &   \textbf{0.59}                                                &  \textbf{0.57}                            \\    \hline                 
\end{tabular}}
						\caption{\small{\textit{Prototype Updating Strategy (\S \ref{subsec:Prototypical})}}}
						\label{table:updating}
\end{subtable}  \ \  
\begin{subtable}{0.42\linewidth}   
						\captionsetup{width=.95\linewidth}
						\resizebox{\textwidth}{!}{
							\setlength\tabcolsep{5.5pt}
							\renewcommand\arraystretch{0.9}
							\begin{tabular}{l|ll|ll}
\hline
\rowcolor[HTML]{EFEFEF} 
\cellcolor[HTML]{EFEFEF}                                   & \multicolumn{2}{c|}{\cellcolor[HTML]{EFEFEF}Flow (val)}                                                & \multicolumn{2}{c}{\cellcolor[HTML]{EFEFEF}Depth (val)}                                               \\
\rowcolor[HTML]{EFEFEF} 
\multirow{-2}{*}{\cellcolor[HTML]{EFEFEF}Head Dimension} & \multicolumn{1}{c}{\cellcolor[HTML]{EFEFEF}Clean} & \multicolumn{1}{c|}{\cellcolor[HTML]{EFEFEF}Final} & \multicolumn{1}{c}{\cellcolor[HTML]{EFEFEF}Clean} & \multicolumn{1}{c}{\cellcolor[HTML]{EFEFEF}Final} \\ \hline  \hline
1                                          &   0.47                                                &          0.76                                          &   0.68                                                &     0.65                                              \\
4                                                  &     0.43                                              &   0.69                                                 &   0.62                                                &    0.61                                               \\
\textcolor{red}{\textbf{8}}                                    &  \textbf{0.40}                                                 &     0.66                                               &                                \textbf{0.59}                   &   0.57                                                \\
16                     &    0.40                                              &                       \textbf{0.65}                             &     0.59                                              &   \textbf{0.55}                           \\    \hline                 
\end{tabular}}
						\caption{\small{\textit{Head Dimension (\S \ref{subsec:ProTer})}}}
						\label{table:head_dim}
\end{subtable}  
\vspace{-3mm}
\caption{\textbf{A set of ablative studies} on Sintel~\cite{butler2012naturalistic} flow and depth datasets (see \S\ref{subsec:exp_ablation}). The adopted designs are marked in {\color{red}red}.}   
\label{table:ablation}
\vspace{-2.5mm}
\end{table*}

\subsection{Ablative Study}
\label{subsec:exp_ablation}
We conduct a series of ablation studies on \textsc{ProMotion} with evaluating on the validation set of Sintel, as in Table~\ref{table:ablation}. 

\noindent\textbf{Key Components.} 
We first assess the effect of each core component in \textsc{ProMotion}. As seen in Table~\ref{table:key}, solely baseline architecture without any proposed designs performs much worse than the model with feature denoiser. With further prototypical learner, we can get lower errors on all datasets, from 0.51 on clean pass to 0.45 on flow. The full settings achieves the best for both flow and depth, which validates the benefit of \textsc{ProMotion}.

\noindent\textbf{Attention Types.} 
We also try vanilla self-attention, multi-head self-attention~\cite{dosovitskiy2020vit}, and spatial reduction attention~\cite{wang2021pyramid} in Table~\ref{table:attention}. Considering that vanilla self-attention is a simple case of multi-head attention with only a single head, and spatial reduction attention focuses only on local context and ignores global dependencies, multi-head self-attention achieves the lowest error among the compared three, but still worse than the subspace self-attention in \textsc{ProMotion}.

\noindent\textbf{Number of Prototypes.} 
Table~\ref{table:nums_proto} reports the performance of our approach with respect to different numbers of prototypes. 
The number of prototypes 10 achieves a baseline performance.
As the number gradually increases to 100, we observe a clear performance gain from 0.46 to 0.40 in error for flow and from 0.65 to 0.59 for depth. However, continuing to increase beyond 100 results in marginal changes.

\noindent\textbf{Decoder Iteration Number.} 
To gain insights into the adaptation head, we ablate the effect of iteration number $T$ in Table~\ref{table:num_iteration}. We find that the performance gradually improves from 0.58 and 0.84 on clean pass for flow and depth to 0.40 and 0.59 when increasing $T$ from 1 to 12, but remains almost unchanged after running more iterations.
However, with more iteration steps, the training and inference time will be stretched. We therefore set $T$ = 12 by default for a better trade-off between accuracy and cost.

\noindent\textbf{Prototype Updating Strategy.} 
We further probe the influence of prototype updating methods, by comparing it with vanilla cosine similarity, K-Means, and Gaussian Mixture Models. As shown in Table~\ref{table:updating}, the adoption is effective, improving cosine similarity and K-Means by 3.3\% and 0.9\%.

\noindent\textbf{Head Dimensions.} 
Finally, we compare the different head dimensions for the attention heads in Table~\ref{table:head_dim}. We find that error significantly reduces from 0.47 to 0.40 in \textit{AEPE} for flow and from 0.68 to 0.59 in $Abs\ Rel$ for depth
when increasing the dimension from 1 to 8, but there is no obvious decrease as the dimension grows to 16.
For a fair comparison with other methods~\cite{huang2022flowformer, dong2023rethinking}, we choose it as 8.

%% file: 10_conclusion.tex
\vspace{-1mm}
\section{Conclusion}
\label{sec:conclusion}
\vspace{-2mm}

In this work, we centered on a prototypical motion paradigm and a feature denoising optimization, which advocates a unified motion framework named \textsc{ProMotion}. This work aims to mitigate the uncertainties by modeling motion as prototypical features and optimizing them in feature space. Empirical results suggest that \textsc{ProMotion} achieves superior performance on two most representative motion tasks: optical flow and scene depth.
Our work may potentially benefit the broader domain of 
dense motion estimation.